\documentclass{article}

\PassOptionsToPackage{numbers, compress}{natbib}

\usepackage[preprint]{neurips_2026}

\usepackage[utf8]{inputenc} 
\usepackage[T1]{fontenc}    
\usepackage{hyperref}       
\usepackage{url}            
\usepackage{booktabs}       
\usepackage{amsfonts}       
\usepackage{nicefrac}       
\usepackage{amsmath}

\usepackage{microtype}      
\usepackage{xcolor}         
\usepackage{graphicx}
\usepackage{algorithm}
\usepackage{algorithmic}
\usepackage{svg}
\usepackage[numbers]{natbib}
\title{Dyna-DINO: Efficient ViT Distillation Via
Adaptive Representation Anchoring}

\author{%
  Jiaqi Zhang$^1$ \hspace{2em} Ashton Lee$^2$ \hspace{2em} Anthony Wong$^1$ \\[1ex]
  \bfseries John Zou$^1$ \hspace{2em} Sami BuGhanem$^1$ \hspace{2em} Randall Balestriero$^1$
}

\begin{document}
\maketitle
\vspace{-0.9cm}
\begin{center}
    \fontsize{9}{8}\selectfont
    \hspace{0.75em} $^1$Brown University \hspace{0.75em}
    $^2$Rice University \hspace{0.75em}\\[1ex]
    \texttt{\{jiaqi\_zhang6, anthony\_g\_wong, john\_zou, sami\_bou\_ghanem, randall\_balestriero\}@brown.edu} \\
    [1ex]\texttt{awl10@rice.edu} \\
\end{center}

\begin{abstract}

Vision Foundation Models (VFMs) with Vision Transformer (ViT) backbones, such as DINOv2, have become essential for downstream tasks like object recognition and semantic segmentation. The immense computational requirements of backbones often necessitate distillation into smaller architectures for edge deployment. Feature-based knowledge distillation (KD) often suffers from the teacher-student gap; the student struggles to imitate teacher's complex feature map due to its limited capacity. To mitigate this bottleneck, we propose \textbf{Dyna-DINO}: Efficient ViT Distillation Via Adaptive Representation Anchoring, a training curriculum for ViT feature-based knowledge distillation. By utilizing the teacher’s intermediate feature maps as a sequence of progressively more difficult targets, our curriculum allows the student to build a foundational representation before tackling higher-level abstractions. Our results demonstrate that this paradigm significantly accelerates convergence through adaptive difficulty selection across various student model sizes and dataset scales. With our curriculum, the Dyna-DINO distilled ViT-S achieves \textbf{$90.1\%$} accuracy on ImageNet-100, a \textbf{$+12.24\%$} improvement compared with baseline. On ImageNet-1K, Dyna-DINO achieves \textbf{$+3.9\%$} and \textbf{$+6.09\%$} improvement for the instance retrieval task on the Oxford and Paris datasets, \textbf{$+1.93\%$} improvements on semantic segmentation task, as well as meaningful performance gain on classification task. Furthermore, the curriculum enables \textbf{$25.1\%$} savings in training FLOPs and \textbf{$21\%$} savings in training time on ImageNet-100 by implementing early-stopping for teacher inference during the initial stages of training. Code is available at \url{https://github.com/KevinZ0217/Dyna-DINO}

\begin{figure}[htbp]
    \centering
    \includegraphics[width=1\textwidth]{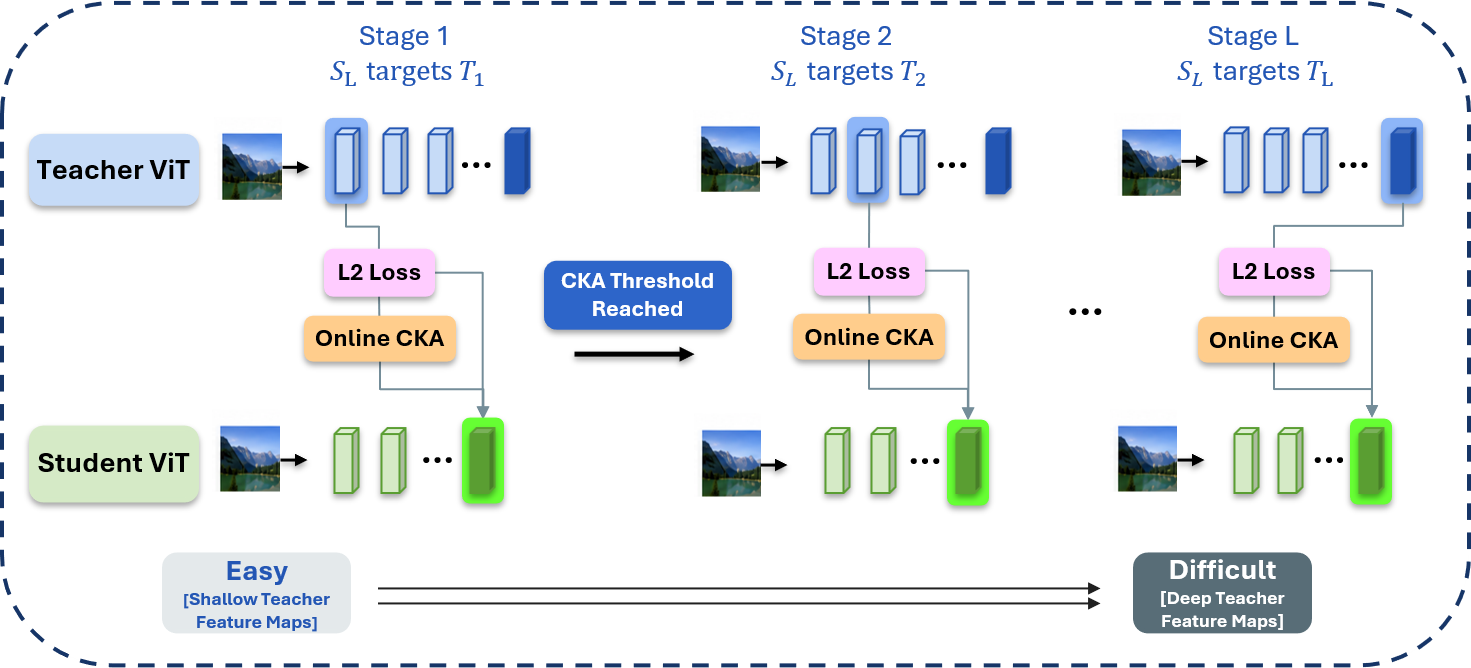}
    \caption{Overview of Dyna. Rather than supervising the student against a fixed teacher block from the start, our curriculum advances the supervisory target through the teacher's feature maps shallow-to-deep based on online CKA alignment, building student representations progressively.}
    \vspace{-13pt}

    \label{fig:teaser}
\end{figure}

\end{abstract}

\section{Introduction}

Vision Transformers (ViT) have transformed computer vision by replacing convolutional hierarchies with spatial self-attention mechanisms that treat image patches as tokens~\cite{dosovitskiy2021image}. While large-scale Vision Foundation Models (VFMs) like DINOv2 achieve state-of-the-art performance, their massive parameter counts—often reaching the scale of ViT-Giant (2B parameters) or ViT-Huge (700M parameters)—make them impractical for deployment on resource-constrained edge devices. Knowledge distillation (KD) is a common solution to distill these "teacher" models into compact "student" models, such as ViT-Small (22m parameters)~\cite{touvron2021training}. Feature-based KD is especially effective for ViTs, as it forces the student to mimic the teacher’s intermediate latent representations rather than just final classification logits~\cite{romero2014fitnets}. This approach allows the distilled student to retain the teacher's versatility across various downstream tasks, including classification, retrieval, and segmentation.

However, a fundamental challenge remains: the teacher-student gap. Research suggests that as the teacher model grows larger, the performance of a small student model often degrades~\cite{guo2023reducing}. Because the student has significantly lower representational capacity and a lower-rank feature space than the teacher. Attempting to match the complex, high-dimensional feature maps of a 40-layer ViT-G teacher in a single step causes unstable training and slow convergence; the student struggles to learn the teacher's final abstractions before it has mastered basic spatial structures.

To bridge this gap, existing literature has explored several strategies to "soften" the distillation target. A primary approach involves intermediate layer matching, where the student is provided with "hints" from the teacher's intermediate layers rather than just the final representation~\cite{hinton2015distilling}. For example, Patient Knowledge Distillation (PKD) introduced fixed-layer selection strategies—such as matching every $k$-th layer (PKD-Skip) or the final $k$ layers (PKD-Last)—to ensure the student captures the hierarchical transformation of information~\cite{sun2019patient} throughout the network. Other frameworks, such as ViTKD~\cite{yang2022vitkd}, use projection heads—often linear or MLP-based layers—to map the student’s intermediate lower-dimensional features into the teacher’s intermediate high-dimensional manifold to perform feature map generation and mimicking, attempting to resolve the physical dimension mismatch and apply hidden supervision. Relational distillation attempts to mitigate this by matching the similarity between patches rather than their raw values~\cite{park2019relational}~\cite{semanticrelation}. Despite these advancements, most existing methods still rely on a static mapping schedule, where the student is forced to align with complex deep-layer features from the very beginning of training, or mapping the student and teacher's intermediate features manually. This "all-at-once" approach fails to account for the student's evolving capacity, and in the scenario where teacher and student have a large size mismatch the layer-mapping becomes arbitrary. 

To mitigate this, we draw inspiration from curriculum learning: a training strategy that introduces concepts in an "easy-to-hard" progression~\cite{bengio2009curriculum}. The similarity analysis reveals that the shallower layers of ViT teacher produce feature maps with a higher similarity score with the final student feature map on earlier training epochs, and the similarity peak gradually shifts to the final teacher feature map as training proceeds. According to this finding, we hypothesize that similar feature maps are easier to learn for the student, thus narrowing the teacher-student gap. By treating the teacher’s shallower, more reconstructive layers as early and accessible targets, and gradually sweeping toward deeper, semantic layers, we guide the student through a structured learning path that accelerates convergence and improves final feature alignment. With this assumption, we propose \textbf{Dyna-DINO}: a training curriculum that gradually switches the training target across intermediate teacher features, with early stopping (controlled by a similarity threshold) pacing the curriculum With experiments on both ImageNet-100 and ImageNet-1K, our distillation curriculum achieves remarkable convergence speed-ups and substantial savings in training FLOPs and wall-time. Evaluations on semantic segmentation, image retrieval and image classification demonstrate that the distilled model retains the ability to adapt to a variety of downstream tasks. 
In summary, our contributions are as follows:

1. We propose \textbf{Dyna-DINO}: a layer-skipping curriculum for feature-based distillation for vision transformers, without the need for manual assignment or intermediate feature selection.

2. A thorough analysis on the robustness for the \textbf{Dyna-DINO} curriculum as well as the effect of using intermediate feature maps to supervise student model's features with knowledge distillation. We furthermore verify and evaluate the  distilled model on image, instance, and pixel level tasks.

\section{Related Work}
\paragraph{Vision Transformer} Vision Transformers (ViTs)~\cite{dosovitskiy2021image} have achieved state-of-the-art performance in a wide array of downstream applications. Due to their exceptional scalability with respect to dataset size and model capacity, ViTs have become the preferred backbone for modern Vision Foundation Models (VFMs), such as DINOv2~\cite{dinov2} and CLIP~\cite{radford2021clip}. A key characteristic of the ViT architecture is in its structural hierarchy: shallower blocks tend to capture local spatial details, whereas deeper layers specialize in extracting dense semantic abstractions~\cite{yang2022vitkd, raghu2021do}. When trained in a self-supervised manner, these models exhibit generalization capabilities that enable high performance in zero-shot or frozen-backbone scenarios, including image classification, semantic segmentation, and instance retrieval, without requiring extensive task-specific fine-tuning.

\paragraph{Knowledge Distillation}While high-capacity ViT backbones offer state-of-the-art performance, their computational requirements necessitate distillation for resource-constrained environments. Knowledge Distillation (KD)~\cite{hinton2015distilling}~\cite{kdsurvey}bridges this gap by transferring knowledge from large-scale teachers to efficient students. Unlike logit-based methods~\cite{sun2024logit} that focus on task-specific output distributions, we utilize feature-based KD~\cite{yang2022vitkd}~\cite{kdcategory}~\cite{ji2021cross}, which minimizes the MSE between student and teacher feature maps via a linear projection layer. This paradigm is more robust and task-agnostic, making it ideal for distilling foundation models where final projection heads may be unavailable. By focusing on intermediate representations, our method ensures the distilled student retains a rich, general-purpose feature space capable of supporting diverse downstream tasks, from object recognition to dense prediction.

\paragraph{Teacher-Student Gap}

Due to the mismatch in capacity between the teacher and student models, a larger teacher model does not necessarily lead to a stronger  distilled student, and, in some cases, can hamper the student model's performance. This phenomenon is referred to as the teacher-student gap~\cite{distilldynamics}. Several prior works propose solutions to reduce the teacher's ability to be more compatible with student's capacity~\cite{jin2019knowledge} , or to address the capacity mismatch by examining the gradient similarity~\cite{yang2023student}. Normalization for the logits has also been shown to be effective for the logit-based distillation~\cite{cho2019efficacy}. Other works indicate that the usage of additional middle-sized teacher models~\cite{mirzadeh2020improved} or intermediate teacher checkpoints during pretraining is also useful for reducing the gap and helpful for distillation into the student~\cite{li2022curriculum}. In summary, the approaches from prior works mainly involve processing the output from the teacher model in ways that are more compatible with the student's capacity. 

\paragraph{Curriculum Learning}Curriculum Learning (CL)~\cite{bengio2009curriculum} improves training by presenting samples in increasing order of difficulty. While CL has recently been used to boost the efficiency of Vision Foundation Models~\cite{fastdinov2, lu2025ditch}, its application to knowledge distillation remains limited. Existing CL-KD strategies, such as logit temperature scaling~\cite{li2022curriculum} or progressive layer matching~\cite{wang2018progressive}, often struggle with feature-based distillation or rely on arbitrary block-wise assignments in situations when teacher and student architectures differ significantly in depth. To address these limitations, we design a curriculum that leverages the teacher’s full range of intermediate features. By ordering these features according to an adaptive similarity metric, our approach eliminates the need for manual layer alignment, facilitating a more natural and efficient transfer of knowledge across disparate architectures.

\section{Recipe for Efficient ViTKD}
\label{headings}

\subsection{Shifted Similarity}
For our baseline, we adopt a standard feature-based KD paradigm that minimizes the $L_2$ distance between the final feature maps of the student and teacher architectures. In this setup, the teacher’s final representation remains the static target throughout the entire training duration. Given that the ViT-G teacher comprises of 40 blocks, whereas the ViT-S student has only 12, a fundamental question arises:

    \textit{Is the final teacher feature map the most accessible target for the student to learn?}

\begin{figure}[htbp]
    \centering
    \includegraphics[width=0.8\textwidth]{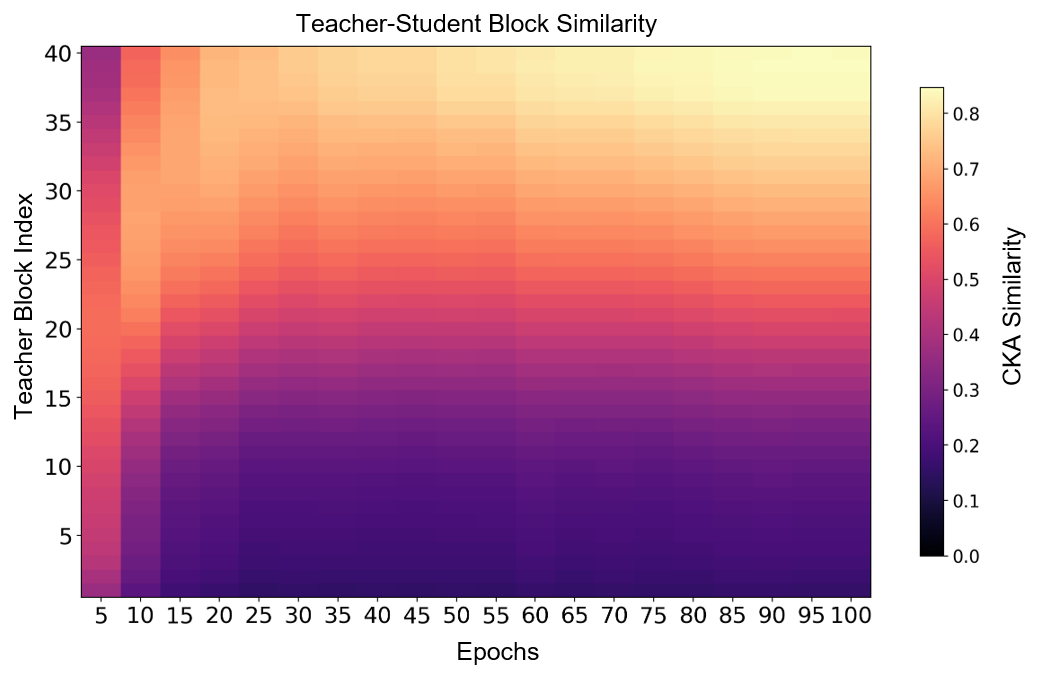}
    \caption{CKA heatmap between the student model's last feature map and all of the teacher's intermediate feature maps during training. Student checkpoints are saved every 5 epochs, and the CKA score is calculated across a subset of validation dataset. }

    \label{fig:cka_heatmap}
    
\end{figure}

To investigate this, we perform a layer-wise similarity analysis between the student’s final feature map and all intermediate feature maps of the teacher during baseline training (see Figure \ref{fig:cka_heatmap}). We utilize Centered Kernel Alignment (CKA) as our metric, calculating scores directly on the raw feature maps without intermediate linear projections.

The resulting similarity landscape reveals that the final teacher layer is not always the most similar to the student's output, despite being the sole training target. Instead, we observe a distinct temporal shift: shallower teacher layers exhibit significantly higher similarity during the initial stages of training, with the peak of this similarity distribution gradually advancing toward the final teacher block as optimization progresses. Operating on the premise that \textit{feature similarity correlates with learning ease}, these findings suggest that a static target is sub-optimal. Rather than focusing on a single layer's representation, the student should adaptively navigate the full range of the teacher’s representational space, dynamically selecting the "easiest" target at each stage of the distillation process.

\vspace{-5pt}

\subsection{The Layer-Skipping Curriculum}

The observed shift in similarity patterns provides the core motivation for our proposed training curriculum. We redefine the complexity of the distillation task by mapping it to the teacher’s depth: shallower teacher feature maps serve as initial, more accessible learning targets, while deeper maps represent increasingly complex semantic abstractions. The training process begins with the student imitating the teacher's first ViT block. Rather than following a fixed, manual schedule, we introduce an adaptive progression mechanism controlled by a CKA similarity threshold ($\tau$). Throughout the distillation process, we monitor the "online" CKA similarity between the student’s terminal feature map and the current teacher target. The curriculum advances to the next teacher block if and only if the similarity score reaches the predefined threshold $\tau$. This similarity-driven transition ensures that the student has sufficiently learned the current level of abstraction before attempting to bridge the next gap in the representational hierarchy. This adaptive logic is formally detailed in Algorithm~\ref{alg:cka_curriculum}.

\section{Experiments}
\subsection{Dataset and Training Setup}
\paragraph{Dataset} 
We evaluate the proposed framework on ImageNet-100~\cite{vinyals2016matching} and the large-scale ImageNet-1K~\cite{russakovsky2015imagenet} datasets. To facilitate rapid experiment iteration, we utilize ImageNet-100 for all ablation studies and parameter tuning. To assess the quality and transferability of the distilled representations, we evaluate performance across several downstream tasks: semantic segmentation is measured on the ADE20K dataset~\cite{zhou2017scene, zhou2019semantic}, while instance retrieval is benchmarked on the revisited Oxford and Paris datasets~\cite{radenovic2018revisiting}. 

\paragraph{Training Setup}

For all distillation experiments, we adopt the standard data augmentation pipeline from~\cite{lightlytrain2025}. Optimization is performed using the LARS optimizer~\cite{you2017large} with a global batch size of 256. The base learning rate is set to 9.0 for ImageNet-100 experiments and 6.0 for ImageNet-1K. Our teacher model utilizes a ViT-G backbone pre-trained with the DINOv2 objective and registers~\cite{dinov2}. For the student models, we evaluate both ViT-Small (ViT-S) and ViT-Tiny (ViT-T), both of which are randomly initialized without registers. Training and evaluation are conducted across a heterogeneous compute cluster featuring NVIDIA L40S, Nvidia RTX A6000, NVIDIA L40, and GeForce RTX 2080 Ti GPUs. All experiments trained on ImageNet-100 cost 12-14 NVIDIA L40S hours, and all ImageNet-1K distillations take 300-400 NVIDIA L40S hours. For all the experiments result tables, we refer "Fixed-anchor" as the baseline for consistently anchoring the student with teacher feature map, "Dyna" as our Dyna-DINO method, and "ViTKD" as another baseline from ~\cite{yang2022vitkd}.

\begin{algorithm}
\caption{CKA-Triggered Progressive Distillation Curriculum}
\label{alg:cka_curriculum}
\begin{algorithmic}[1]
    \REQUIRE Teacher features $\{T_1, \dots, T_M\}$, student $S$, threshold $\tau$, patience $E_{max}$
    \STATE $m \gets 1$, \quad $e \gets 0$ \COMMENT{current target block, epochs on it}
    \FOR{epoch $= 1$ \TO $N_{epochs}$}
        \STATE Sample batch $B$
        \STATE $score \gets \mathrm{CKA}\bigl(S_{last}(B),\, T_m(B)\bigr)$
        \IF{$m < M$ \textbf{and} ($score \geq \tau$ \textbf{or} $e \geq E_{max}$)}
            \STATE $m \gets m + 1$, \quad $e \gets 0$ \COMMENT{advance curriculum}
        \ELSE
            \STATE $e \gets e + 1$
        \ENDIF
        \STATE $\mathcal{L} \gets \mathrm{MSE}\bigl(S_{last}(B),\, T_m(B)\bigr)$
        \STATE Update $S$ via $\nabla \mathcal{L}$
    \ENDFOR
    \RETURN $S$
\end{algorithmic}
\end{algorithm}
\subsection{The Layer-Skipping Curriculum Saves Training Time and FLOPs}
We first validate the effectiveness of our curriculum on ImageNet-100. The distillation objective is defined as:
\begin{equation}
    \mathcal{L}_{\mathrm{distill}} = \mathrm{MSE}\left( P(S_{\mathrm{feat}}), T_{\mathrm{feat}} \right) + 0.05 \cdot \mathrm{MSE}\left( P(S_{\mathrm{cls}}), T_{\mathrm{cls}} \right)
\end{equation}
Following \cite{zhou2022ibot}, we utilize a single-layer linear projector $P$ to align the student’s hidden dimensions with those of the teacher; this projector is shared between the patch tokens and the CLS token. We maintain a constant weight of 0.05 for the CLS token loss throughout all experiments to ensure a consistent comparison.

\begin{figure}[htbp]
    \centering
    \includegraphics[width=1\textwidth]{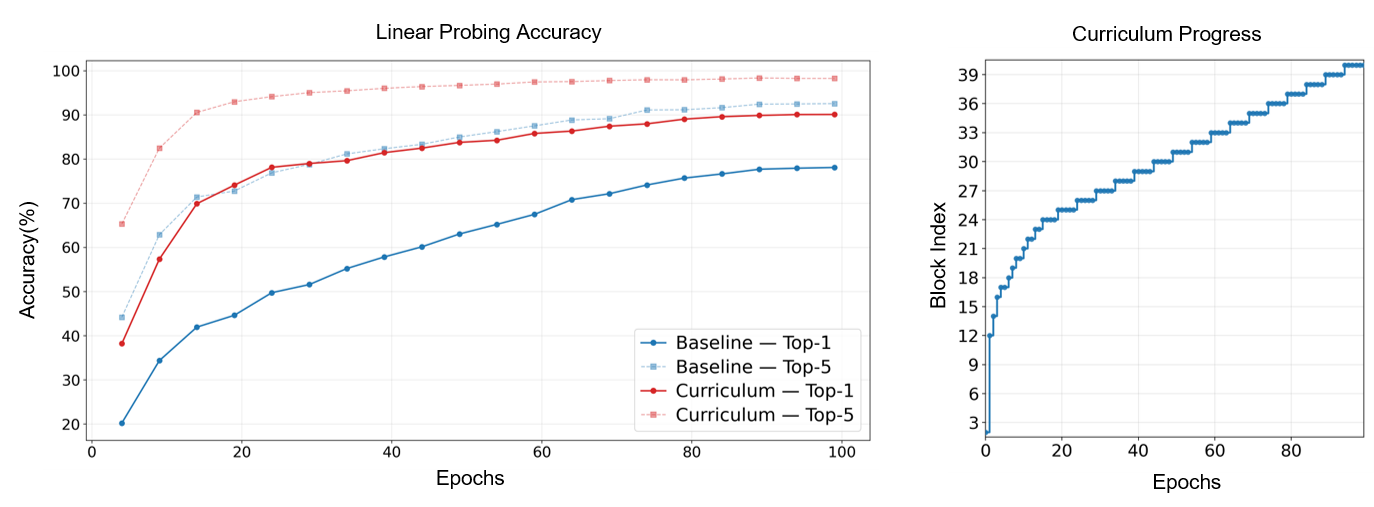}
    \vspace{-13pt}
    \caption{Left: the linear probing accuracy convergence comparison for baseline and Dyna. Right: the curriculum progress visualization. The student checkpoint for linear probing evaluation is saved for every 5 epochs.}
    \label{fig:linear_probing_curriculum_pace}
\end{figure}

We evaluate convergence speed by performing linear probing on student checkpoints every five epochs. As illustrated in Figure \ref{fig:linear_probing_curriculum_pace} (left), Dyna distillation demonstrates significantly faster convergence than the baseline from the earliest stages of training. This acceleration suggests that shallower teacher feature maps effectively serve as "easier" learning targets, allowing the student to establish a stable representational foundation before tackling more complex objectives.

The progression of the curriculum, visualized in Figure \ref{fig:linear_probing_curriculum_pace} (right), reveals a distinct temporal pattern. The student transitions rapidly through initial teacher blocks which primarily encode local spatial information, while dwelling significantly longer in the deeper layers. This behavior suggests that deeper layers contain denser semantic knowledge that needs more steps to learn. This evidence validates our assumption of depth serving as a reliable proxy for target difficulty.

To evaluate the generalizability of our curriculum, we investigate its performance across different student capacities: ViT-Tiny (12 blocks, $\sim$6M parameters) and ViT-Small (12 blocoks, $\sim$22M parameters). As detailed in Table \ref{tab:vit_comparison_in100}, Dyna consistently enhances linear probing accuracy for both architectures, with ViT-S reaching 90.10\% and ViT-Tiny achieving 81.76\%. Critically, these performance gains are accompanied by significant computational savings; the adaptive early-stopping for teacher inference achieves up to a 28.8\% reduction in training FLOPs and a 22.5\% reduction in total training time.

\begin{table}[t!]
    \centering
    \caption{Performance and efficiency comparison between baseline distillation and Dyna. Students (ViT-S, ViT-Tiny) are distilled from a ViT-G teacher on ImageNet-100 for 100 epochs. Dyna utilizes a CKA-threshold of 0.85. Efficiency metrics (FLOPs and Train Time) represent the reduction in teacher computational overhead during the distillation process.}
    \label{tab:vit_comparison_in100}
    \begin{tabular}{lccccc}
        \toprule
        \textbf{Experiments} & \textbf{Lin. Probe} $\uparrow$ & \textbf{Lin. Probe} $\uparrow$ & \textbf{FLOPs} & \textbf{Train Time} \\
        & (patch token) & (CLS token) & \textbf{Saving} & \textbf{Saving} \\
        \midrule
        Fixed-anchor (ViT-S)     & 77.86\% & 77.12\% & --- & --- \\
        ViTKD (ViT-S)~\citep{yang2022vitkd} & 70.74\% & 66.88\%  & --- & --- \\
        Fixed-anchor (ViT-Tiny)  & 75.90\% & 76.04\%  & --- & --- \\
        \midrule
        Dyna (ViT-S)   & \textbf{90.10}\% & \textbf{89.66}\%  & 25.1\% & 21\% \\
        Dyna (ViT-Tiny) & \textbf{81.76}\% & \textbf{82.02}\%  & 28.82\% & 22.5\% \\
        \bottomrule
    \end{tabular}
\end{table}

\vspace{-5pt}

\subsection{Evaluating Representational Quality via Downstream Tasks}

While the accelerated convergence of the proposed curriculum is demonstrated through image-level linear probing, we further evaluate the distilled models on diverse downstream tasks to assess the generalizability of the learned representations. We conduct evaluations on semantic segmentation and instance retrieval to verify that the student model preserves both pixel-level details and fine-grained instance features. These experiments confirm that our curriculum-based distillation not only speeds up training but also produces a versatile backbone capable of adapting to a wide range of applications.

To evaluate instance-level representations, we perform image retrieval experiments on the Oxford and Paris datasets. For each image, the model backbone extracts a global embedding; we then compute the similarity scores between the query image embedding and the database embeddings. These results are ranked by similarity to calculate the mean Average Precision (mAP). As shown in Table~\ref{tab:oxford_paris_results_with_mean}, our Dyna-distilled ViT-S and ViT-Tiny outperform the baseline by a significant margin across all three difficulty levels. These results demonstrate that the proposed curriculum effectively preserves the fine-grained structural details required for precise instance matching.

We evaluate performance on ADE20K using three protocols: Linear Segmentation, Encoder-Only Mask Transformer (EOMT), and Multi-Scale (MS) inference. The results are summarized in Table~\ref{tab:segmentation_results_100}.

In the Linear Segmentation setting, which measures the linear separability of frozen features, Dyna improves the ViT-S mIoU from 12.15\% to 20.53\%. When utilizing the EOMT head to evaluate how features interact within a Transformer-native decoder, the performance gap remains significant, with the Dyna-distilled ViT-S reaching 38.10\% mIoU compared to the baseline's 24.49\%. Finally, under Multi-Scale (MS) evaluation (a protocol that tests robustness to scale variations) the Dyna-distilled ViT-S achieves 39.36\% mIoU, representing a 14.74\% absolute improvement. These consistent gains across different decoding architectures and inference scales suggest that navigating the teacher's representational hierarchy during training allows the student to retain denser spatial and semantic information than standard distillation regimes.

\begin{table}[t!]
    \centering
    \caption{Instance recognition performance comparison (mAP) on Oxford and Paris datasets. The "Mean" column represents the average mAP across Easy, Medium, and Hard difficulty levels.}
    \label{tab:oxford_paris_results_with_mean}
    \begin{tabular}{lccccccccc} 
        \toprule
        & \multicolumn{4}{c}{roxford5k} & & \multicolumn{4}{c}{rparis6k} \\
        \cmidrule{2-5} \cmidrule{7-10}
        Training method & E & M & H & \textbf{Mean} & & E & M & H & \textbf{Mean} \\
        \midrule
        Fixed-anchor (ViT-S)      & 10.28 & 8.80  & 2.15 & 7.08  & & 24.83 & 21.29 & 6.63  & 17.79 \\
        ViTKD (ViT-S)      & 7.48 & 6.82  & 1.37 & 5.22  & & 28.55 & 21.20 & 7.25  & 18.79 \\
        Fixed-anchor (ViT-Tiny)   & 8.04  & 7.72  & 1.84 & 5.87  & & 21.35 & 19.42 & 6.59  & 15.79 \\
        \midrule
        Dyna (ViT-S)          & \textbf{22.56} & \textbf{17.30} & \textbf{4.81} & \textbf{14.89} & & \textbf{53.84} & \textbf{40.99} & \textbf{15.97} & \textbf{36.93} \\
        Dyna (ViT-Tiny)       & \textbf{14.70} & \textbf{12.15} & \textbf{2.84} & \textbf{9.90}  & & \textbf{43.39} & \textbf{33.36} & \textbf{12.38} & \textbf{29.71} \\
        \bottomrule
    \end{tabular}
\end{table}

\begin{table}[htbp]
    \centering
    \caption{Semantic segmentation performance comparison for distillation on ImageNet-100. Results are reported in mean Intersection over Union (mIoU) for linear segmentation, EOMT validation, and multi-scale (MS) evaluation.}
    \label{tab:segmentation_results_100}
    \begin{tabular}{lccc}
        \toprule
        \textbf{Method} & \textbf{Linear Seg.} $\uparrow$ & \textbf{EOMT val} $\uparrow$ & \textbf{MS} $\uparrow$ \\
        & (mIoU) & (mIoU) & (mIoU) \\
        \midrule
        Fixed-anchor (ViT-S)       & 12.15\% & 24.49\% & 24.62\% \\
        ViTKD (ViT-S) ~\citep{yang2022vitkd}       & 11.51\% & 14.94\% & 14.75\% \\
        Fixed-anchor (ViT-Tiny)    & 9.51\% & 15.41\% & 14.88\% \\
        \midrule
        Dyna (ViT-S)     & \textbf{20.53}\% & \textbf{38.10}\% & \textbf{39.36}\% \\
        Dyna (ViT-Tiny)  & \textbf{14.71\%} & \textbf{21.67}\% & \textbf{21.82\%} \\
        \bottomrule
    \end{tabular}
    
\end{table}

\subsection{Effectiveness on Large Dataset}
Following the demonstrated effectiveness on ImageNet-100, we evaluate the scalability of the Dyna curriculum on ImageNet-1K. We introduce a more challenging baseline; in addition to only anchor-
ing the student’s final feature map (which we refer to as Fixed-anchor(1 Anchor), we anchor the
student’s 1st, 6th, and the final feature map and name this baseline Fixed-anchor (3 Anchors)

The advantages of Dyna remain evident in tasks requiring high-fidelity features. As shown in Table \ref{tab:oxford_paris_results_1k}, Dyna achieves remarkable performance across the Oxford and Paris retrieval benchmarks. Specifically, the Dyna-distilled ViT-S achieves a mean mAP improvement of 3.9\% on the Oxford dataset and 6.09\% on the Paris dataset. These results suggest that even at scale, Dyna effectively transfers the structural nuances which are critical for fine-grained instance matching.

In semantic segmentation (Table \ref{tab:segmentation_results_1k}) and linear probing (Table \ref{tab:vit_comparison_in1k}), we observe that Dyna produces competitive results with the standard distillation baseline. In semantic segmentation, the Dyna-Distilled ViT-S shows significantly better performance on linear segmentation, EOMT, and Multi-scale evaluation. Similarly, linear probing accuracies on ImageNet-1K demonstrate the effectiveness with ViT-S reaching 78.10\% (vs. 77.63\% and 72.31\% baseline) and ViT-Tiny reaching 64.62\% (vs. 64.4\% baseline).
We have an interesting observation that while fixed-anchor(3 anchors) obtains better performance than fixed-anchor(1 anchors) on image retrieval task, it performs worse on the segmentation and linear probing, which require high linear separability in the underlying representation. Dyna outperforms both baselines on both classification and retrieval tasks, and we attribute this benefit to the fact that Dyna curriculum effectively distills a more versatile and functional representation for downstream tasks. 

While ViTKD is able to show a matching performance on ImageNet-100 experiments, it does not converge well in the ImageNet-1K scale training. In the original paper of ViTKD, all the distillation experiments are conducted with teacher/student with similar parameter size(ViT-B teacher with ViT-S student, for example.) We attribute the result to this design choice: the method is designed for teacher and student with similar capability, and when it comes to more challenging scenario with huge teacher-student gap (ViT-G teacher and ViT-S student), the assumption that teacher and students are structural similar no longer stand, and this method could fail. 

While achieving competitive performance on global classification and dense prediction tasks, Dyna provides a measured improvement in training efficiency, yielding a 12.4\% reduction in teacher FLOPs and a 7.5\% decrease in training time for the ViT-Tiny student. Though these computational savings are modest at the ImageNet-1K scale, they are achieved while enhancing the model's feature integrity. Specifically, the substantial gains observed in the image retrieval benchmarks, where Dyna outperforms the baseline by up to 6.09\%, serve as a strong proxy for overall representation quality. The results confirm that Dyna produces a feature space that better preserves the complex representation of the foundation model teacher.

\begin{table}[t!]
    \centering
    \caption{Results for ImageNet-1K. Performance comparison (mAP) on Oxford and Paris datasets. The "Mean" column represents the average mAP across Easy (E), Medium (M), and Hard (H) difficulty levels.}
    \label{tab:oxford_paris_results_1k}
    \begin{tabular}{lccccccccc}
        \toprule
        & \multicolumn{4}{c}{roxford5k} & & \multicolumn{4}{c}{rparis6k} \\
        \cmidrule{2-5} \cmidrule{7-10}
        Training method & E & M & H & \textbf{Mean} & & E & M & H & \textbf{Mean} \\
        \midrule
        Fixed-anchor (S, 1An)      & 27.34 & 19.36 & 4.86 & 17.19 & & 62.70 & 50.46 & 27.80 & 46.99 \\
        Fixed-anchor (S, 3An)      & 33.85 & 23.31 & 4.64 & 20.6 & & 58.62 & 47.25 & 24.35 & 43.41 \\
        ViTKD (ViT-S)     & 3.87 & 3.88 &   0.87 &2.87  &  & 13.85& 13.07& 5.32 & 10.75    \\
        Fixed-anchor (T, 1An)   & 19.41  & 14.16  & \textbf{2.42}  & 11.9  & & 50.18  &  38.47 & 14.98  &  34.54 \\
        \midrule
        Dyna (S, 3An)    & \textbf{34.12} & \textbf{24.31} & \textbf{4.86} & \textbf{21.1} & & \textbf{68.04} & \textbf{55.78} & \textbf{33.54} & \textbf{52.45} \\
        Dyna (T, 3An) & \textbf{21.93}  & \textbf{14.99}  & 2.40  & \textbf{13.11}  & & \textbf{55.91}  & \textbf{42.97}  & \textbf{17.50}  & \textbf{35.94}  \\
        \bottomrule
    \end{tabular}
\end{table}

\begin{table}[t!]
    \centering
    \caption{Semantic segmentation performance comparison. Results are reported in mean Intersection over Union (mIoU) for linear segmentation, EOMT validation, and multi-scale (MS) evaluation.}
    \label{tab:segmentation_results_1k}
    \begin{tabular}{lccc}
        \toprule
        \textbf{Method} & \textbf{Linear Seg.} $\uparrow$ & \textbf{EOMT val} $\uparrow$ & \textbf{MS} $\uparrow$ \\
        & (mIoU) & (mIoU) & (mIoU) \\
        \midrule
        Fixed-anchor (ViT-S, 1 Anchor)       & 29.85\% & 46.65\% & 48.22\% \\
        Fixed-anchor (ViT-S, 3 Anchors)       & 27.41\% & 36.23\% & 37.20\% \\
        ViTKD (ViT-S) ~\citep{yang2022vitkd}       & 4.28\% & 6.3\% & 6.19\% \\
        Fixed-anchor (ViT-Tiny)    & 18.92\% & 33.93\% & 33.42\% \\
        
        \midrule
        Dyna (ViT-S, 3 Anchors)     & \textbf{31.78}\% & \textbf{48.68\%} & \textbf{50.05\%} \\
        Dyna (ViT-Tiny, 3 Anchors)  & \textbf{20.23\% }& \textbf{37.37\%} & \textbf{37.61\%} \\
        \bottomrule
    \end{tabular}
    
\end{table}

\begin{table}[t!]
    \centering
    \caption{Results for training on ImageNet-1K. Comparison of Baselines and Dyna methods for ViT-S and ViT-Tiny with ViT-G teacher.  for both ViT-Tiny and ViT-S the CKA threshold for curriculum is 0.8. Efficiency metrics (FLOPs and Train Time saving) represent the reduction in teacher computational overhead during the distillation process.}
    \label{tab:vit_comparison_in1k}
    \begin{tabular}{lccccc}
        \toprule
        \textbf{Experiments} & \textbf{Lin. Probe} $\uparrow$ & \textbf{Lin. Probe} $\uparrow$ & \textbf{FLOPs} & \textbf{Train Time} \\
        & (patch token) & (CLS token)  & \textbf{Saving} & \textbf{Saving} \\
        \midrule
        Fixed-anchor (ViT-S, 1 Anchor)     & 76.87\% & 77.63\% & --- & --- \\
        Fixed-anchor (ViT-S, 3 Anchors)     & 73.37\% & 72.31\% &  --- & --- \\
        ViTKD (ViT-S) ~\citep{yang2022vitkd}     & 15.28\% & 12.09\% &  --- & --- \\
        Fixed-anchor (ViT-Tiny, 1 Anchor)  & 63.35\% & 64.4\%  & --- & --- \\
        \midrule
        Dyna (ViT-S, 3 Anchors)   &  \textbf{77.03\%} &  \textbf{78.10\%} &  \textbf{9.44\%} &\textbf{ 4.83\%} \\
        Dyna (ViT-Tiny, 3 Anchors) & \textbf{63.37\%} & \textbf{64.62\%} & \textbf{12.4\%} & \textbf{7.5\%} \\
        \bottomrule
    \end{tabular}
\end{table}

\subsection{The Necessity of Progressive Supervision vs. Single-Layer Targets}
A potential critique of intermediate feature distillation is that the performance gains may stem from selecting a specific "optimal" layer rather than the curriculum itself. If a single intermediate layer contained the most transferable knowledge, static distillation from that "lucky" layer would outperform a progressive curriculum. To investigate this, we conducted an ablation study using a ViT-S teacher and a ViT-Tiny student. We trained twelve separate student models, each supervised exclusively by a different fixed intermediate teacher layer throughout training. This allows us to isolate the contribution of individual layers and determine whether Dyna’s effectiveness is derived from a specific static target or the structured transition between them.

\begin{figure}[htbp]
    \centering
    \includegraphics[width=1\textwidth]{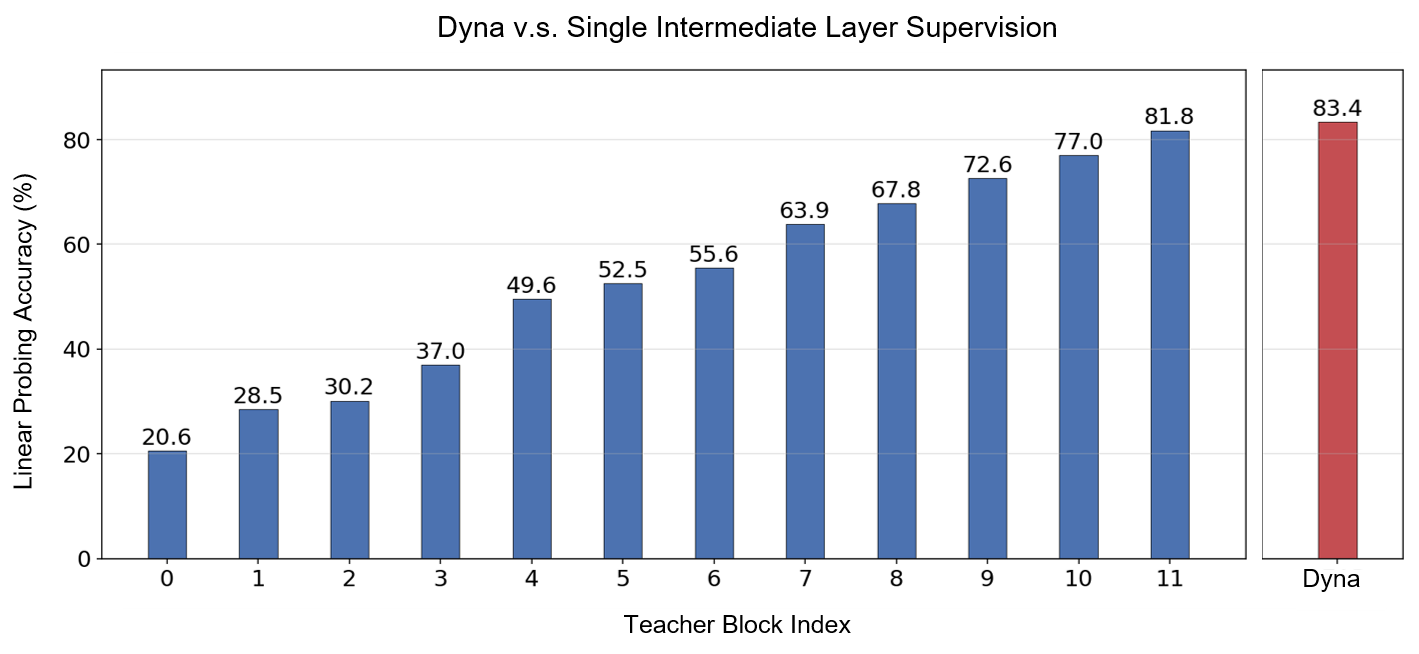}
    \vspace{-13pt}
    \caption{Linear probing accuracy comparison between Dyna and single intermediate layer supervision. Dyna outperforms using any single layer as supervision throughout training, indicating the effectiveness of utilizing a structural curriculum.}
    \label{fig:intermediate_layer_comparison}
\end{figure}

As illustrated in Figure \ref{fig:intermediate_layer_comparison}, while deeper teacher layers generally results in higher feature quality, Dyna outperforms all individual targets. These results suggest that no single frozen intermediate layer is sufficient to bridge the gap between teacher and student as effectively as a curriculum. Instead, the performance benefit is a product of the structured progression itself, which allows the student to learn a hierarchical foundation that any single-layer target lacks.

\subsection{Comparison with Dense One-to-One Layer Alignment}
When teacher and student architectures share an identical depth, a common strategy for maximizing supervision is dense intermediate matching: aligning every student block with its corresponding teacher block, an approach often viewed as the the upper-bound for feature-based distillation. 

In this section, we compare Dyna against this dense matching baseline. We conduct this experiment using a ViT-S teacher/ ViT-Tiny student and ViT-B teacher/ViT-S student; since both models consist of 12 layers, they allow for a direct, one-to-one feature map alignment. The results in table~\ref{tab:alignment_comparison} reveal the efficiency of Dyna; with only 1 projector, Dyna is merely $0.02\%$ behind the dense layer alignment upper bound, which uses 12 projectors in total. In the case of ViT-B teacher with ViT-S student, Dyna even outperforms the dense one to one alignment by 2.86\%. This comparison indicates that a structured, adaptive curriculum can achieve or even surpass dense supervision while maintaining much higher structural simplicity.

\begin{table}[htbp]
    \centering
    \caption{Comparison of distillation alignment strategies. One-to-one alignment refers to matching each teacher layer with corresponding student layer, while Dyna utilizes an adaptive progression curriculum. The Fixed-anchor method aligns the last teacher and student feature map.}
    \label{tab:alignment_comparison}
    \begin{tabular}{lcc}
        \toprule
        \textbf{Alignment Strategy} & \textbf{Linear Probing} $\uparrow$ & \textbf{Projectors (Params)} \\
        \midrule
        Fixed-anchor (S$\rightarrow$ T)  & 81.80\% & $\times 1$ (0.07M) \\ 
        One-to-one Alignment (S$\rightarrow$ T)       & 83.38\% & $\times 12$ (0.89M) \\
        Dyna (S$\rightarrow$ T)  & 83.36\% & $\times 1$ (0.07M) \\
        \midrule
        Fixed-anchor (B$\rightarrow$ S)  & 72.52\% & $\times 1$ (0.3M) \\ 
        One-to-one Alignment (B$\rightarrow$ S)       & 83.82\% & $\times 12$ (3.55M) \\
        Dyna (B$\rightarrow$ S)               & 86.68\% & $\times 1$ (0.3M) \\
        \bottomrule
        
    \end{tabular}
\end{table}

\section{Conclusion, Limitations, and Future Directions}
\label{sec:limitations}
In this work, we propose a layer-skipping curriculum for efficient feature-based knowledge distillation in Vision Transformers (ViTs). Through an automatic similarity-based curriculum, the student model achieves accelerated convergence and learns high-quality representations for diverse downstream tasks. One limitation of our work is the assumption of a white-box teacher model with accessible intermediate features. While many open-weight Vision Foundation Models (VFMs) exist, our method is restricted in scenarios where only the final teacher feature map is available. Furthermore, the CKA threshold utilized for our ImageNet-1K experiments may represent a sub-optimal hyperparameter. A more exhaustive optimization of this threshold could yield further performance improvements. A potential future direction is to expand this framework to cross-architecture scenarios where the teacher and student have different architectures (e.g., Transformer to CNN). Another key direction involves extending our method to distilling across data modalities for multimodal models, such as audio, language, and more.

\bibliographystyle{plainnat}
\bibliography{references}
\appendix

\section{Technical appendices and supplementary material}

\subsection{Linear Probing with Standard Deviation}
\begin{figure}[htbp]
    \centering
    \includegraphics[width=0.7\textwidth]{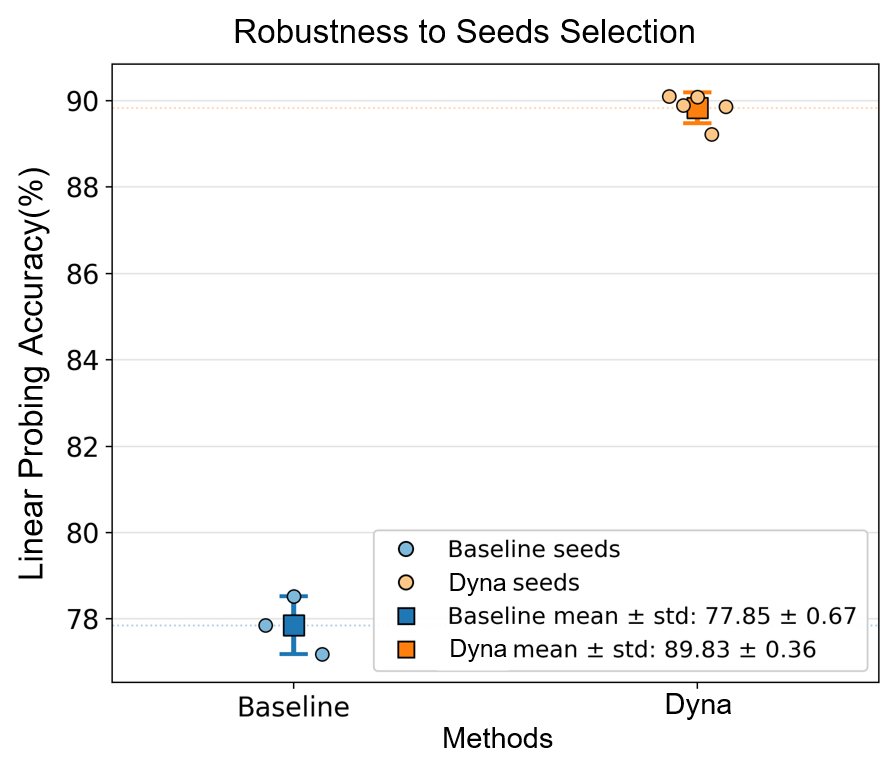}
    \vspace{-13pt}
    \caption{Linear probing accuracy comparison between Dyna and baseline with multiple seeds. Dyna consistently outperforms baseline on this ImageNet-100 distillation regardless of the seeds.}
    \label{fig:robustness_seed}
\end{figure}

In this section, we investigate whether Dyna can perform consistently. ViT-G teacher is used to distill ViT-S student on ImageNet-100, and we selected 0.85 as the CKA threshold for Dyna. 3 distinct seeds are selected for baseline distillation while 5 seeds are used for Dyna distillation. As in figure ~\ref{fig:robustness_seed}, Dyna shows consistent performance across multiple seeds with small standard deviation, indicating the effectiveness of this approach.

\subsection{Curriculum Robustness to CKA Threshold Selection}

\begin{figure}[htbp]
    \centering
    \includegraphics[width=0.7\textwidth]{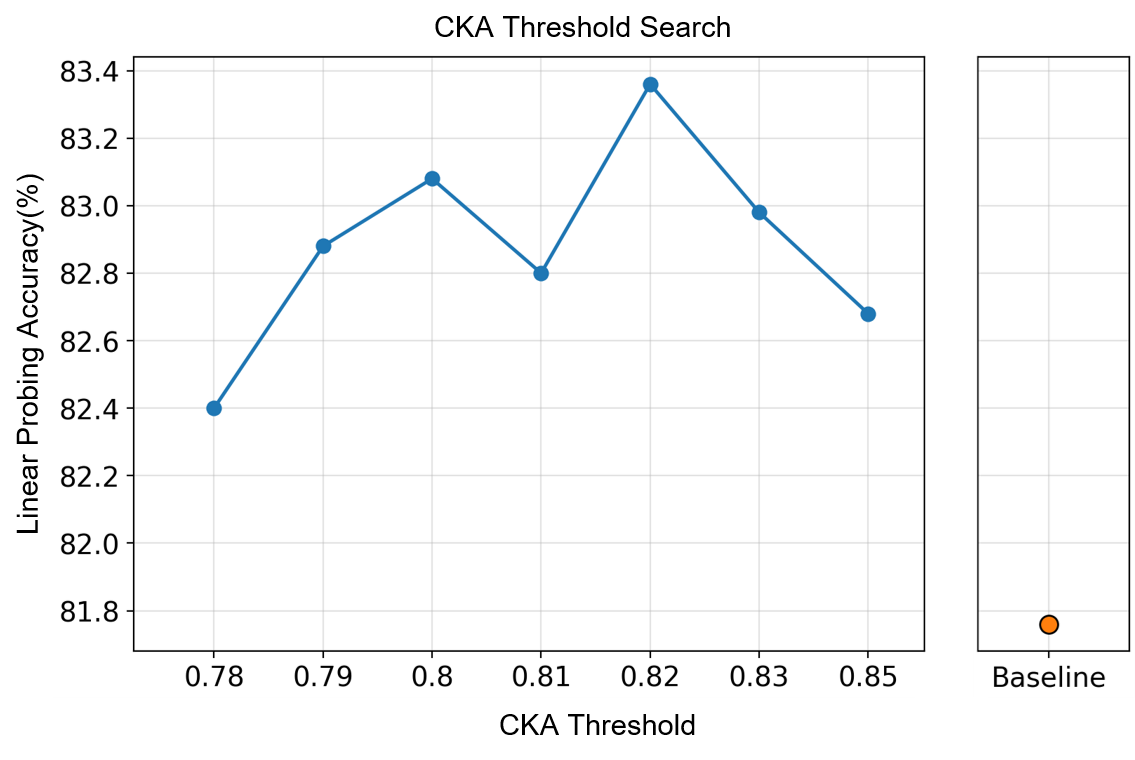}
    \vspace{-13pt}
    \caption{Dyna performance comparison for multiple CKA thresholds. While Dyna is robust to the threshold selections, 0.82 is the optimal threshold for this setting.}
    \label{fig:threshold_search}
\end{figure}
We conduct a small scale cka threshold search for the distillation from ViT-S teacher distilled from DINOv2 and ViT-Tiny student initialized from scratch. As in figure ~\ref{fig:threshold_search}, Dyna is generally robust to multiple threshold selection, and the optimal threshold is around 0.82. 

\end{document}